# Performance Evaluation of Deep Learning Networks for Semantic Segmentation of Traffic Stereo-Pair Images

Vlad Taran, Nikita Gordienko, Yuriy Kochura, Yuri Gordienko, Alexandr Rokovyi, Oleg Alienin, Sergii Stirenko

*National Technical University of Ukraine "Igor Sikorsky Kyiv Polytechnic Institute", Kyiv, Ukraine*

Semantic image segmentation is one the most demanding task, especially for analysis of traffic conditions for self-driving cars. Here the results of application of several deep learning architectures (PSPNet and ICNet) for semantic image segmentation of traffic stereo-pair images are presented. The images from Cityscapes dataset and custom urban images were analyzed as to the segmentation accuracy and image inference time. For the models pre-trained on Cityscapes dataset, the inference time was equal in the limits of standard deviation, but the segmentation accuracy was different for various cities and stereo channels even. The distributions of accuracy (mean intersection over union — mIoU) values for each city and channel are asymmetric, long-tailed, and have many extreme outliers, especially for PSPNet network in comparison to ICNet network. Some statistical properties of these distributions (skewness, kurtosis) allow us to distinguish these two networks and open the question about relations between architecture of deep learning networks and statistical distribution of the predicted results (mIoU here). The results obtained demonstrated the different sensitivity of these networks to: (1) the local street view peculiarities in different cities that should be taken into account during the targeted fine tuning the models before their practical applications, (2) the right and left data channels in stereo-pairs. For both networks, the difference in the predicted results (mIoU here) for the right and left data channels in stereo-pairs is out of the limits of statistical error in relation to mIoU values. It means that the traffic stereo pairs can be effectively used not only for depth calculations (as it is usually used), but also as an additional data channel that can provide much more information about scene objects than simple duplication of the same street view images.

## 1  INTRODUCTION AND BACKGROIUND

With the development of machine learning, in particular deep neural networks, new methods for obtaining information from images of different content appear. The semantic image segmentation is one of the most fundamental and complicated tasks in computer vision. It allows to predict labels for all pixels in the image for deep understanding of scene as a whole, availability of classes of objects (and instances even), their locations, and spatial limits. During the last years new architectures of neural networks appeared, which introduced new improvements to achieve greater accuracy of results for this problems [1]. For example, AlexNet [2], VGG-16 [3], GoogLeNet [4] and ResNet [5] became well-known standard and building blocks for new architectures. Recently, interest to deep learning networks and especially to Fully Convolutional Networks (FCN) increased in the context of image semantic segmentation tasks [1]. The goal of that approach is to take advantage of existing convolutional neural networks (CNNs) to learn hierarchies of features. For this purpose, researchers tried to transform the aforementioned models into fully convolutional ones by replacing fully connected layers with convolutional ones to output spatial maps instead of classification scores. Recently, several new FCN networks, like SegNet [6], ENet [7], PSPNet [8], ICNet [9], DeepLab [10], ResNet [5], and many others, demonstrated the high performance with regard to accuracy and speed of prediction. Usually, performance of these networks is tested against some popular datasets of traffic road conditions like Cityscapes [11] and their results can be found elsewhere (see for example, https://www.cityscapes-dataset.com/benchmarks).

The current tendency is to monitor street scenes by multiple input modalities, for example: by a stereo camera rig in streets from 50 different cities in Cityscapes public dataset [11]; stereo camera, traffic light camera, localization camera [12]; and many others including Tesla Autopilot proprietary solution (https://www.tesla.com/autopilot) with 8 surround cameras in addition to 12 ultrasonic sensors and forward-facing radar; the radar can see vehicles through heavy rain, fog or dust. Using the input data from multiple input modalities, several approaches to process them were applied, for example, stereo datasets (which are of greatest interest for us now) mostly used to get depth information, for example, for semantic segmentation [13-14].

In the view of these background results, the main aim of this paper is to present results of application of several deep learning architectures for semantic image segmentation of traffic stereo-pair images from the public Cityscapes dataset [11] with a quantitative characterization of the prediction results for the left and right channels (parts) of stereo-pairs.

## 2   EXPERIMENTAL AND COMPUTATIONAL DETAILS

For this investigation the images from Cityscapes dataset were used (Fig.1,a-c) that correspond to several German cities like Frankfurt (267 images), Munster (174 images), and Lindau (59 images), which were used for validation in Cityscapes dataset. The wider test Cityscape datasets are under work now and will reported separately. The ground truth images labeled for the left channel images were also used for the right channel images to compare the actual difference between them. Also some custom street view images were taken from the city of Kyiv (Ukraine) (Fig.1,d) just for visual for comparison to give immediate impression that variations between local conditions can be much crucial for cities from different countries. But these results are under work yet and will be reported in details elsewhere.

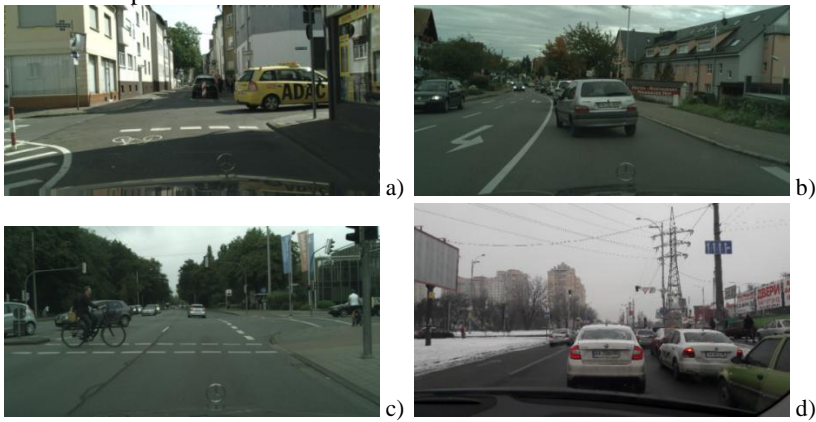

Fig. 1. Examples of images from Cityscapes database: (a) Frankfurt, (b) Lindau, (c) Munster; and custom database: Kyiv (d).

FCN-based PSPNet [8] network and its compressed high-speed version ICNet [9] were selected for semantic image segmentation of traffic stereo-pair images as representatives of the same family of FCN-based networks targeted for semantic image segmentation. All experiments were conducted on the basis of TensorFlow platform on the workstation with the single NVIDIA Titan 1080 GPU card with CUDA 8.0 and CUDNN 7.

## 3   RESULTS AND DISCUSSION

Both PSPNet and ICNet perform the semantic segmentation of the selected original images (Fig.2a and Fig.2c) of traffic conditions and provide their segmented versions (Fig.2b and Fig.2d). The obtained segmented images have the different quality of predicted classes which can obviously seen by the naked eye even (Fig.2), especially for the custom street view images that were not present in Cityscapes dataset (these results will be reported later elsewhere).

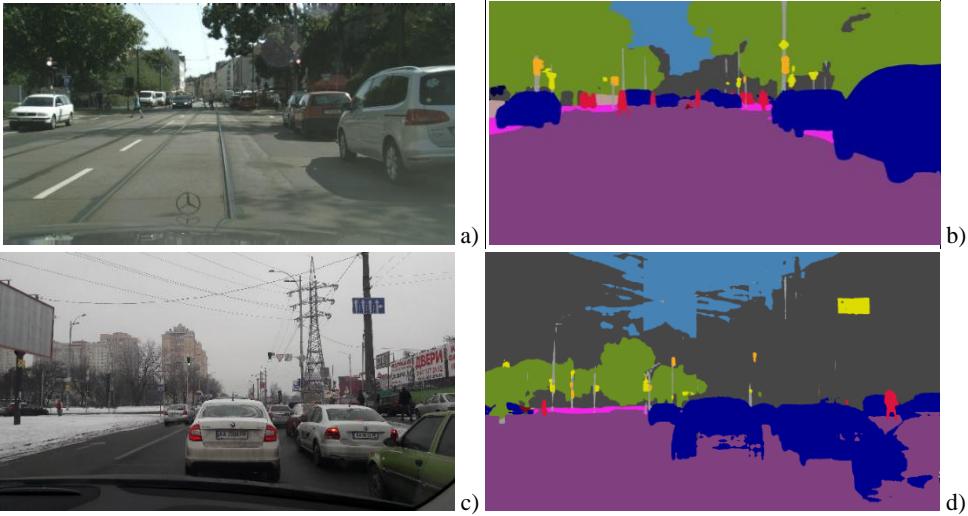

Fig. 2. Original images (a, c) and their segmented versions (b, d) obtained by PSPNet network from Cityscapes (a) and custom (c) images.

Accuracy (mean intersection over union — mIoU) and inference time were used to compare performance of PSPNet and ICNet networks for different cities and channels of stereo-pair.

### 3.1   Accuracy

The accuracy of semantic segmentation measured by mIoU parameter was determined for all images from the mentioned subset of Cityscapes dataset. The distributions of mIoU values were determined for each city (Frankfurt, Munster, Lindau), each channel (left and right), and as a total (Fig.3). Then they were analyzed with regard to their basic distribution parameters: mean, standard deviation, skewness, and kurtosis.

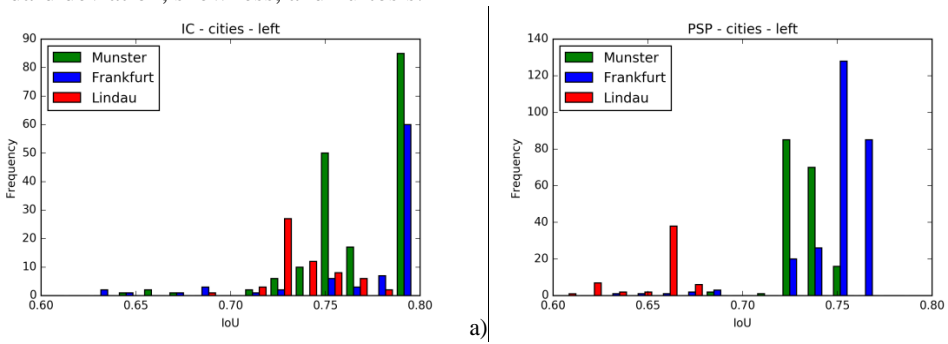

Fig. 3. Distributions of accuracy values (mIoU) for the set of left images obtained by: (a) ICNet network and (b) PSPNet network.

The qualitative analysis of these distributions allows to make assumptions that they are asymmetric, long-tailed, and have many extreme outliers. The quantitative statistical analysis of prediction distributions is used here that recently proved to be effective to find the differences and change of behavior in various ensembles of objects in different applications [15-16]. Below the results of the quantitative analysis of these distributions are given by means of estimation of mean and standard deviation (Fig.4), skewness and kurtosis (Fig.5).

The mean mIoU values for right and left sets of images are quite different for both ICNet and PSPNet networks, and this difference is much bigger than standard deviation for each set (Fig.4).

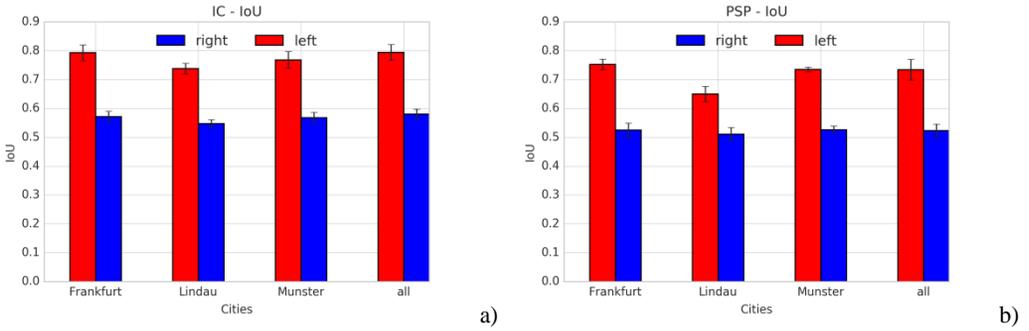

Fig. 4. Mean and standard deviations of accuracy (mIoU) obtained by: (a) ICNet and (b) PSPNet network.

The skewness and kurtosis (Fig.5) are different for both ICNet and PSPNet networks. The distributions of accuracy (mIoU) values for various cities obtained by ICNet network are much closer to the normal distribution than the distributions of accuracy (mIoU) values for various cities obtained by PSPNet. It can be observed by the relative positions of the skewness and kurtosis values in relation to the location of the normal distribution (Fig.5; black square).

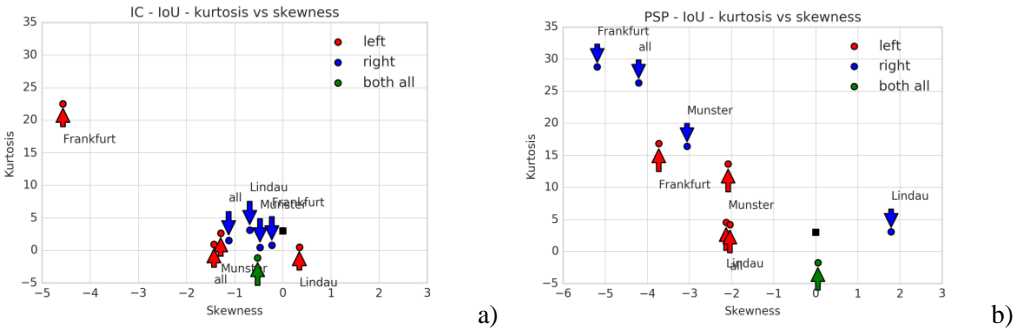

Fig. 5. Skewness and kurtosis of accuracy (mIoU) obtained by: (a) ICNet network and (b) PSPNet network. The green arrow shows the skewness and kurtosis for the set including all (left and right) images for all cities. The black square denotes the place of the normal distribution.

### 3.2 Inference time

The similar analysis was applied for the inference time determined for all images from the mentioned subsets of Cityscapes dataset and the distributions of the inference time values (Fig.6) were analyzed in the same way (Fig.7-8).

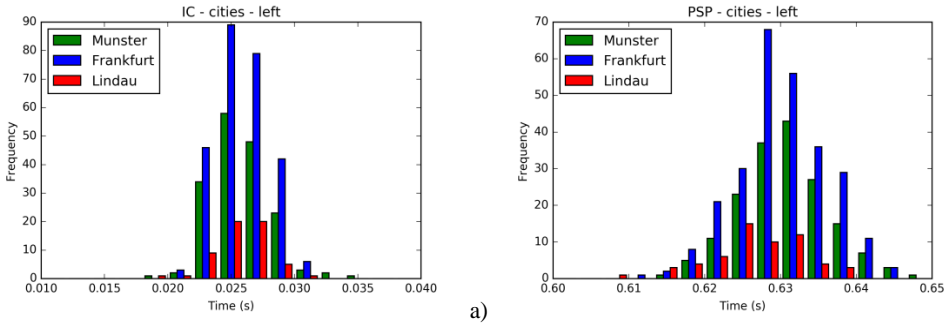

Fig. 6. Distributions of inference times obtained by: (a) ICNet network and (b) PSPNet network.

The qualitative analysis of these distributions allows to make assumptions that they are close to symmetric, short-tailed, and have fewer outliers. Below the results of the quantitative analysis of these distributions are given by means of estimation of mean and standard deviation (Fig.7), skewness and kurtosis (Fig.8).

The mean inference time values for right and left sets of images are not very different for both ICNet and PSPNet networks, and this difference is much lower than standard deviation for each set (Fig.7).

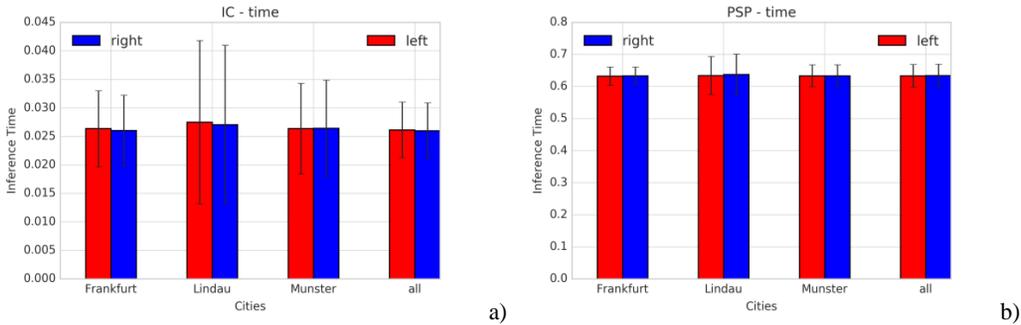

Fig. 7. Time of inference obtained by: (a) ICNet network and (b) PSPNet network.

The skewness and kurtosis (Fig.8) demonstrate the similar tendency for both ICNet and PSPNet networks. The total distribution of the inference time values (Fig.8; green circle) for all cities is close to the normal distribution (Fig.8; black square) for both ICNet and PSPNet networks. But the distributions for each of the cities demonstrate the positive skewness values that is characteristic for the right-skewed distributions where the right-tail corresponds to the bigger inference times. The skewness and kurtosis for right and left sets of images are not very different for both ICNet and PSPNet networks.

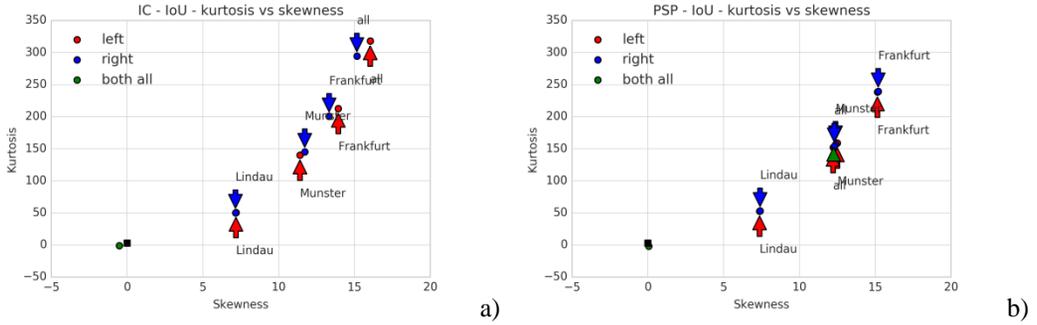

Fig. 8. Time of inference obtained by: (a) IC network and (b) PSPNet network. The green circle shows the skewness and kurtosis for the set including all (left and right) images for all cities. The black square denotes the place of the normal distribution.

### 3.2 Comparison of mIoU and inference time

The combined plots mIoU vs. inference times (Fig.9) allowed us to make comparison of these values with regard to the network applied (ICNet and PSPNet) and the input channel (left and right). For ICNet, the mean mIoU values for all left images (Fig.9; red) are quite different in comparison to mean mIoU for all right (Fig.9; blue) images. And the mean mIoU value for the total distribution (including images from all cities and all channels) is closer to the mean mIoU value for the left input channel. But for PSPNet, the mean mIoU values are not so different for all left (Fig.9; red) and all right (Fig.9; blue) images. And the mean mIoU value for the total distribution is located between the mean mIoU values for the left and right input channels.

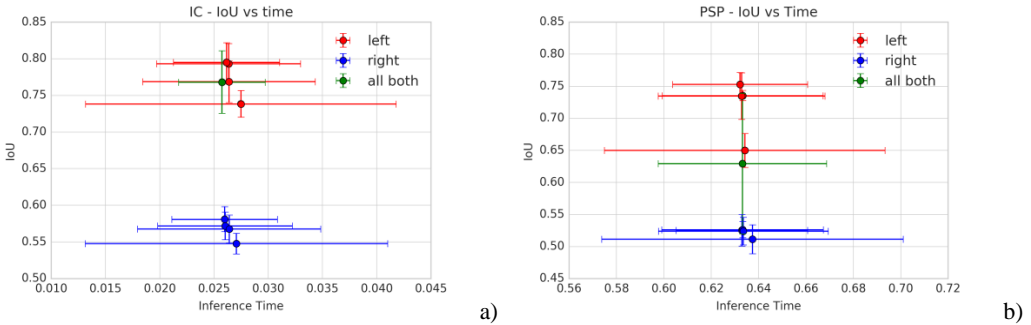

Fig. 9. Accuracy and time of inference obtained by: (a) ICNet network and (b) PSPNet network. Each symbol corresponds to some city (without labels here).

The common plot for both networks (Fig.10) allows to compare their relative position with regard to accuracy (mIoU) and inference time with taking into account the standard deviation of these values for city subsets. Variations of mIoU among cities can be very big, for example, for left images the mIoU obtained by PSPNet for Lindau was equal to 0.65±0.03, while for Frankfurt it was equal to 0.75±0.02. The difference of mIoU values between left and right input channels is evidently big for the both networks also. For example, the mean mIoU obtained by ICNet was equal to 0.80±0.03 for all left images and 0.58±0.02 for all right images, but the mean mIoU obtained by PSPNet was equal to 0.74±0.04 for all left images and 0.52±0.02 for all right images. In this context, information from the additional right data channel is radically different from the point of view of both networks, because it is out of the limits of statistical error.

As to the inference time, ICNet is more stable than PSPNet, because ICNet demonstrates the much narrower (by 10 times) scatter of the inference time values with the total mean 0.026 (left green circle) and standard deviation 0.004 (left green error bars) in comparison to PSPNet with the total mean 0.63 (right green circle) and standard deviation 0.04 (right green error bars).

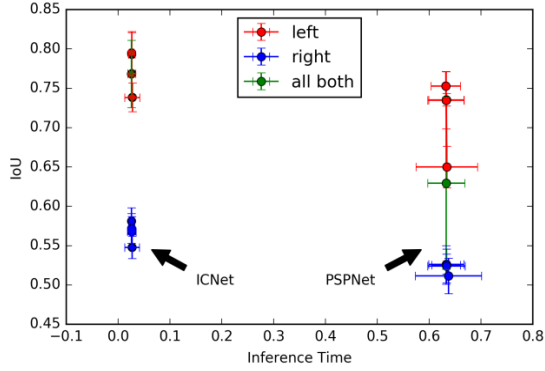

Fig. 10. Accuracy (mIoU) versus inference time obtained by: (a) ICNet network and (b) PSPNet network.

## 4   CONCLUSIONS

In summary, we have carried out the performance tests of PSPNet an ICNet against several subsets of Cityscapes dataset including stereo-pair images taken by left and right cameras for different cities. It was found that the distributions of mIoU values for each city and channel are asymmetric, long-tailed, and have many extreme outliers, especially for PSPNet network in comparison to ICNet network. To the moment the reasons of these statistical properties are unclear yet, but they allow us to distinguish these two networks and open the question about relation between architecture of deep learning networks and statistical distribution of the predicted results (mIoU here). The results obtained demonstrated the different sensitivity of these networks to: (1) the local street view peculiarities (among different cities), (2) the change of viewing angle on the same street view image (right and left data channels). The differences with regard to the local street view peculiarities should be taken into account during the targeted fine tuning the models before their practical applications. For both networks, information from the additional right data channel is radically different from the left channel, because it is out of the limits of statistical error in relation to mIoU values. It means that the traffic stereo pairs can be effectively used not only for depth calculations (as it is usually used), but also as an additional data channel that can provide much more information about scene objects than simple duplication of the same street view images.

We believe that this work can stimulate more interest to relation between architectures of deep learning networks and statistical distributions of the predicted results, and also to the deeper investigation of the multichannel data extension in deep learning tasks, especially for other uses of the traffic stereo-pair images for semantic segmentation of traffic conditions, and in more applications like robotics, online face/gesture recognition, symbol/writing recognition [17], and even wearable computing in the context of elderly care [18].